\documentclass[conference]{IEEEtran}
\IEEEoverridecommandlockouts
\usepackage{cite}
\usepackage{amsmath,amssymb,amsfonts}
\usepackage{algorithmic}
\usepackage{graphicx}
\usepackage{textcomp}
\usepackage{xcolor}
\def\BibTeX{{\rm B\kern-.05em{\sc i\kern-.025em b}\kern-.08em
    T\kern-.1667em\lower.7ex\hbox{E}\kern-.125emX}}

\begin{document}
\newpage
\onecolumn
\noindent
{\Huge\textbf{\\\\\\\\IEEE Copyright Notice}}\\\\\\\\

\noindent
{\Large \textcopyright~2020 IEEE.  Personal use of this material is permitted.  Permission from IEEE must be obtained for all other uses, in any current or future media, including reprinting/republishing this material for advertising or promotional purposes, creating new collective works, for resale or redistribution to servers or lists, or reuse of any copyrighted component of this work in other works.\\\\\\}

\noindent
{\Large Accepted to be published in: Proceedings of the IEEE Design Automation Conference (DAC), 2020.}

\twocolumn
\newpage
\title{Efficient Synthesis of Compact Deep Neural Networks}\vspace{5em}
\author{\IEEEauthorblockN{Wenhan Xia}
\IEEEauthorblockA{
\textit{Princeton University}\\
wxia@princeton.edu}
\and
\IEEEauthorblockN{Hongxu Yin}
\IEEEauthorblockA{
\textit{Princeton University}\\
hongxuy@princeton.edu}
\and
\IEEEauthorblockN{Niraj K. Jha}
\IEEEauthorblockA{
\textit{Princeton University}\\
jha@princeton.edu}}
\maketitle

\begin{abstract}
Deep neural networks (DNNs) have been deployed in myriad machine learning 
applications. However, advances in their accuracy are often achieved with 
increasingly complex and deep network architectures. These large, deep models 
are often unsuitable for real-world applications, due to their massive 
computational cost, high memory bandwidth, and long latency. For example, 
autonomous driving requires fast inference based on Internet-of-Things (IoT) 
edge devices operating under run-time energy and memory storage constraints. 
In such cases, compact DNNs can facilitate deployment due to their reduced 
energy consumption, memory requirement, and inference latency. Long short-term 
memories (LSTMs) are a type of recurrent neural network that have also found 
widespread use in the context of sequential data modeling. They also face a 
model size vs. accuracy trade-off. In this paper, we review major approaches 
for automatically synthesizing compact, yet accurate, DNN/LSTM models suitable 
for real-world applications.  We also outline some challenges and future areas 
of exploration.
\end{abstract}


\section{Introduction}
Deep neural networks (DNNs) have revolutionized various artificial 
intelligence applications, such as computer vision, speech recognition, 
machine translation, and smart healthcare~\cite{human_performance, speechlstm, seq2seq, diabdeep}. Unlike traditional machine learning methods that process hand-crafted features extracted from raw data, DNNs automatically learn to represent data features via multi-level abstraction. Today, DNNs even outperform humans in some tasks, such as image classification~\cite{he2016deep}. 

To achieve high accuracy, researchers tend to design wider and deeper
DNN models. This requires massive computational resources and immense
amounts of data for training~\cite{krizhevsky2012imagenet,
simonyan2014very, he2016deep}. The slowdown in Moore's Law, however,
makes it difficult to keep pace with the high memory bandwidth and
computational power requirements of these increasingly large models.
Consequently, there is a widening gap between computational demands from
DNNs and resources/capabilities offered by the underlying hardware, thus
limiting DNN deployment in many real-world applications. For example,
the substantial storage, memory bandwidth, and energy requirements may
be too high for edge devices, such as mobile phones, smart watches, and
Internet-of-Things (IoT) sensors~\cite{him, akmandor2018smart}. Bulky
DNNs also tend to have a high inference latency, which is unacceptable
in various delay-sensitive scenarios, such as autonomous
driving~\cite{wu2018squeezeseg}. In addition to computational power demands, the immense labeled datasets needed to train these models can be prohibitively costly and time-consuming to obtain. Dataset acquisition also raises privacy concerns, especially in domains like 
healthcare, where patient data must be protected~\cite{yin2017health}. 

LSTMs are a type of neural network that incorporate feedback. This
enables them to perform sequential data modeling. They find use in
applications like speech recognition \cite{deepspeech2}, neural machine
translation~\cite{seq2seq}, health monitoring~\cite{deepheart}, and
language modeling~\cite{stanford,wenwei}. LSTM accuracy is typically
increased by increasing its depth, hence its size. Obtaining a compact, yet accurate, LSTM is also a challenging task.

In this paper, we provide an overview of effective techniques to address the above problems and facilitate DNN/LSTM deployment across the IoT hierarchy -- from the cloud to edge to sensor.

The rest of this paper is organized as follows. In Section~\ref{sec:overview}, we review DNN development history and deployment constraints, and summarize existing approaches for designing compact models. Section~\ref{sec:efficiency} focuses on major techniques that yield accurate, yet compact, models for fast inference. Section~\ref{sec:HA} describes compact DNN designs that take into account hardware specifications and traits. Section~\ref{sec:data-efficiency} discusses data constraints and availability, and techniques for overcoming these limitations. Finally, 
Section~\ref{sec:summary} provides a summary, and outlines some research challenges for future exploration.  

\begin{figure*}[t]
\centerline{\includegraphics[width=14cm]{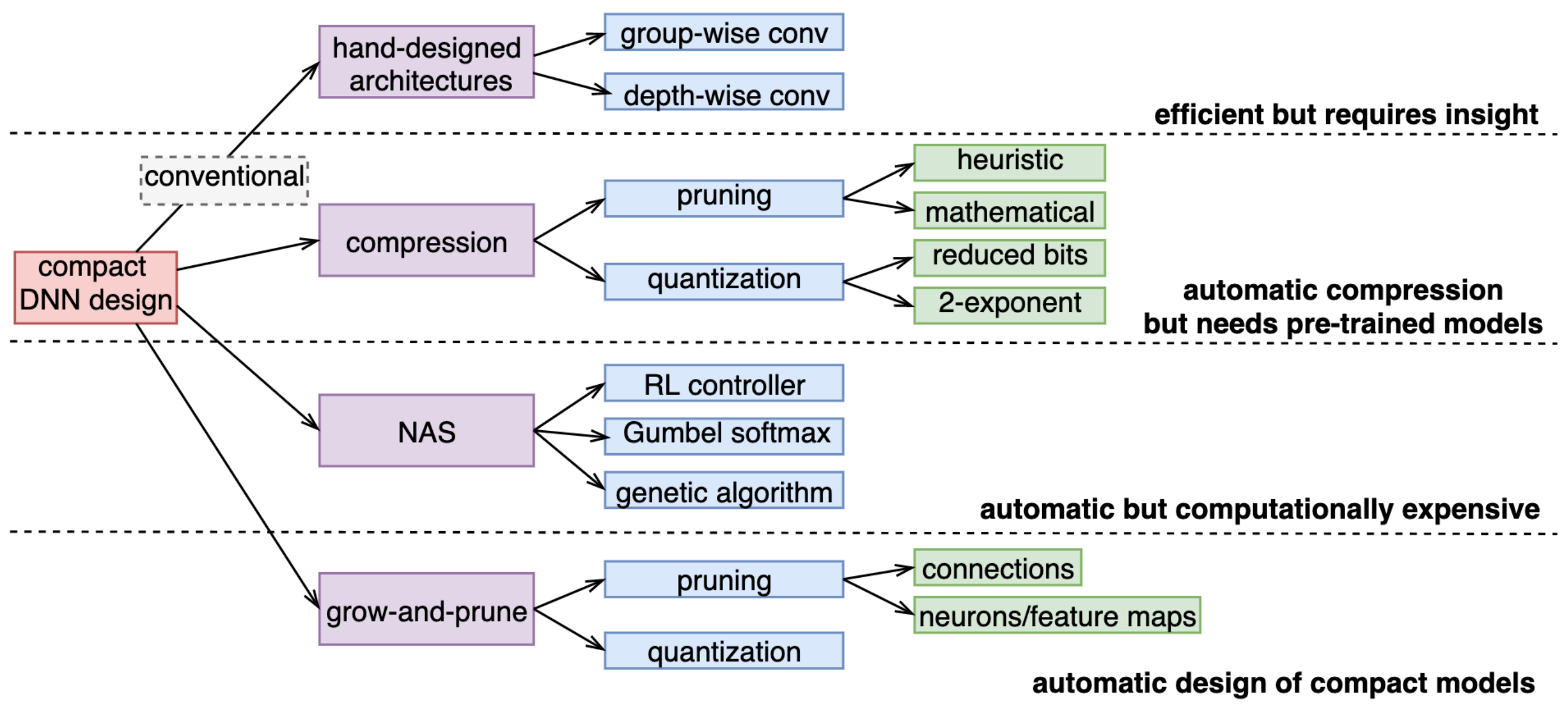}}
\caption{Overview of approaches for compact DNN design; 
RL: reinforcement learning; NAS: neural architecture search.}
\vspace{-2mm}
\label{fig:main}
\end{figure*}

\section{Overview}~\label{sec:overview}
Deep learning is a branch of  machine learning that uses DNNs, which learn high-level data representations through many processing layers. Due to its outstanding and robust performance, deep learning has been adopted by many academic disciplines, from engineering to biology to psychology.  

\subsection{A Brief History of DNNs}
Although the popularity of deep learning only bloomed in the past decade, its foundations were laid in the 1940s, when McCulloch and Pitts showed that networks of artificial neurons, later referred to as neural networks, could execute first-order logical functions\cite{mcculloch1943logical}. In 1958, Rosenblatt developed the perceptron that performed threshold-based classification on numerical inputs \cite{rosenblatt1958perceptron}. Following Rosenblatt's perceptron, Ivakhnenko and Lapa developed a multilayer
perceptron: one of the first multi-layer neural network architectures \cite{ivakhnenko1966cybernetic}. In the l990s, LeCun et al. proposed LeNet-5\cite{lecun1998gradient}, a neural network trained with the backpropagation algorithm. LeNet-5 is commonly regarded as the first practical application of neural networks and was widely used to recognize hand-written digits on checks. 

The deep learning revolution occurred in the early 2010s. It is often
attributed to three major factors: availability of large-scale datasets,
significantly increased computational speed provided by GPUs, and novel DNN architectures, especially those involving convolutional layers. The ImageNet Large Scale Visual Recognition Challenge (ILSVRC)~\cite{deng2009imagenet} 
embodies the rapid evolution of deep learning during this decade. In 2012, AlexNet~\cite{krizhevsky2012imagenet}, a deep convolutional neural network (CNN), soundly defeated traditional machine learning methods. AlexNet inspired active research and development of deep CNNs, including VGG~\cite{simonyan2014very} and GoogLeNet~\cite{szegedy2015going}. In 2015, another deep learning milestone was established when the deep residual network, ResNet~\cite{he2016deep}, achieved superhuman top-5 classification accuracy. In addition to breaking the human performance barrier, ResNet also solved the problem of stacking more layers with a residual framework. The rapid development of deep learning is mirrored by the substantial decrease in the top-5 classification error every year.

\subsection{Complications}
The yearly improving accuracy on the ImageNet dataset is associated with increasingly deeper DNN models. For example, AlexNet used eight layers, VGG 16 layers, GoogLeNet 22 layers, and ResNet 152 layers. A similar trend is observable in the field of speech recognition. For example, DeepSpeech2~\cite{deepspeech2}, which has been widely used for speech recognition, is more than two times deeper and ten times larger than the 
initial DeepSpeech.

Most modern DNNs are deep and bulky, which may be an appropriate design
choice when given enough training time, data, and computational resources. However, this ideal scenario does not necessarily hold for many real-world applications. We categorize constraints on developing DNNs into the following four groups: 

\begin{itemize}
\item \textbf{Computational energy}:  Inference platforms, such as wearable devices and autonomous drones, are highly energy-constrained, whereas DNN models are expected to continually make predictions. Thus, designing DNNs that can satisfy stringent energy budgets is now a major design objective.

\item \textbf{Latency}: Applications like autonomous driving, real-time video analysis, and speech processing assistants like Siri and Cortana require fast inference. This limits the number of parameters and layers that a DNN can have. 

\item \textbf{Memory}: Deploying large, highly parameterized DNNs requires 
massive memory usage. This is expensive and infeasible for many applications. 

\item \textbf{Data availability}:
Obtaining large-scale labeled training datasets can be costly and
time-consuming. Most available datasets are of small or medium size. In
addition, certain datasets, e.g., biomedical, may not be publicly
available due to privacy concerns. Hence, designing DNNs that can
achieve high performance in a limited data regime is a major new
research thrust. 
\end{itemize}

\subsection{Overview of Compact DNN Design}
Due to the limitations described above, considerable effort is being invested in efficient DNN design. Next, we summarize the major approaches for developing compact DNN models, as illustrated in Fig.~\ref{fig:main}. 

Conventional approaches for building efficient DNNs involve the creation
of efficient building blocks for removing redundancy. For example,
MobileNetV2~\cite{sandler2018mobilenetv2} stacks inverted residual
building blocks to shrink model size. Ma et al. achieve model
compactness via channel shuffle operations and depth-wise
convolution~\cite{ma2018shufflenet}. Wu et al. propose ShiftNet, a
framework for using a shift-based module rather than spatial convolutional
layers, to substantially reduce computational and storage cost~\cite{wu2017shift}. Although these hand-crafted models can offer impressive compactness, they require substantial design insight and trial-and-error modifications to maximize performance. 

Network compression is an alternative, automatic approach for compact
DNN design that eliminates the need for {\em a priori} design principles
and trial-and-error modification. It is easy to implement on pre-trained
models and generally requires less development time. Pruning is a
widely-used compression technique that removes individual weights or
entire filters (neurons) such that model performance, e.g., accuracy, 
does not drop significantly. An ideal pruning solution would be to compute the $\ell_0$ norm of weights. However, this is non-convex, NP-hard, and requires combinatorial search. These issues can be avoided with heuristics-based pruning criteria that are strongly correlated with the ground truth importance estimates of weights or neurons for final inference~\cite{molchanov2016pruning, thinet, nisp, molchanov2019importance}. For example, Han et al. have shown the effectiveness of weight pruning based on magnitude and achieved substantial non-regular weight sparsity in modern CNNs~\cite{han2015deep}. Subsequent work has extended the methodology to $\ell_1$ norm~\cite{li2016pruning}, Taylor expansion~\cite{molchanov2016pruning, molchanov2019importance}, and batch norm~\cite{liu2017learning, ye2018rethinking} based filter pruning for achieving structural sparsity. The model can be further compressed through low-bit quantization. For example, Zhu et al. show that CNNs with a low-bit representation of weights require significantly less memory for image recognition and object detection tasks, with only a slight loss in accuracy~\cite{zhu2016trained}. The Hardware-Aware Automated Quantization framework leverages reinforcement learning (RL) to automatically determine the quantization policy by incorporating hardware accelerator feedback in the design loop~\cite{wang2019haq}.

Recently, ideas from RL have been adapted to search for DNN architectures in an automated flow. This approach is known as neural architecture search 
(NAS)~\cite{rl3, nasnet}. NAS typically uses a controller, e.g., a recurrent neural network (RNN), to iteratively generate groups of candidate neural networks in the search process. Candidate performance is later used as a reward for enhancing the controller~\cite{NASRL}. A recent work implements these ideas with an architecture called NASNet that achieves better performance than hand-crafted DNNs by using RL to search for architectural building blocks~\cite{nasnet}. Furthermore, RL-based NAS can also be used to develop efficient DNNs for mobile devices and platforms. For example, MnasNet~\cite{tan2018mnasnet} running on a Pixel phone achieves 75.2\% top-1 accuracy on the ImageNet classification dataset with only 78ms latency. 

Despite their success in compact architecture search, RL-based methods remain computationally intensive. To enable an effective and direct gradient descent in the architecture space, Wu et al. propose FBNet that utilizes Gumbel softmax to jointly optimize connections and weights using the same objective function~\cite{fbnet}. Genetic algorithms offer an alternative approach for this optimization. For example, a methodology called NEAT simultaneously learns neural network topologies and weights, resulting in optimized and increasingly complex models over generations~\cite{neat}. However, it is beneficial to employ extremely efficient search algorithms based on various predictors in this context, as explained later in the discussion of the ChamNet approach~\cite{chamnet}.

NAS and related methods are appropriate when a large number of GPUs are available for training. However, this is not the case in many scenarios. Grow-and-prune offers an alternative approach for deriving efficient models 
without the need for many GPUs. In addition, unlike network compression, grow-and-prune does not require a pre-trained model. It simultaneously learns both network weights and architecture. A grow-and-prune synthesis paradigm typically reduces the number of parameters in multi-layer perceptron, CNN, and LSTM based models by another 2$\times$ relative to when only pruning is used, while increasing accuracy~\cite{dai2019nest,dai2018grow, hassantabar2019scann}.

\section{Model Efficiency}
\label{sec:efficiency}
In this section, we discuss synthesis tools that achieve both model efficiency and high accuracy. 

\begin{figure}[t]
\begin{center}
\includegraphics[width=\columnwidth]{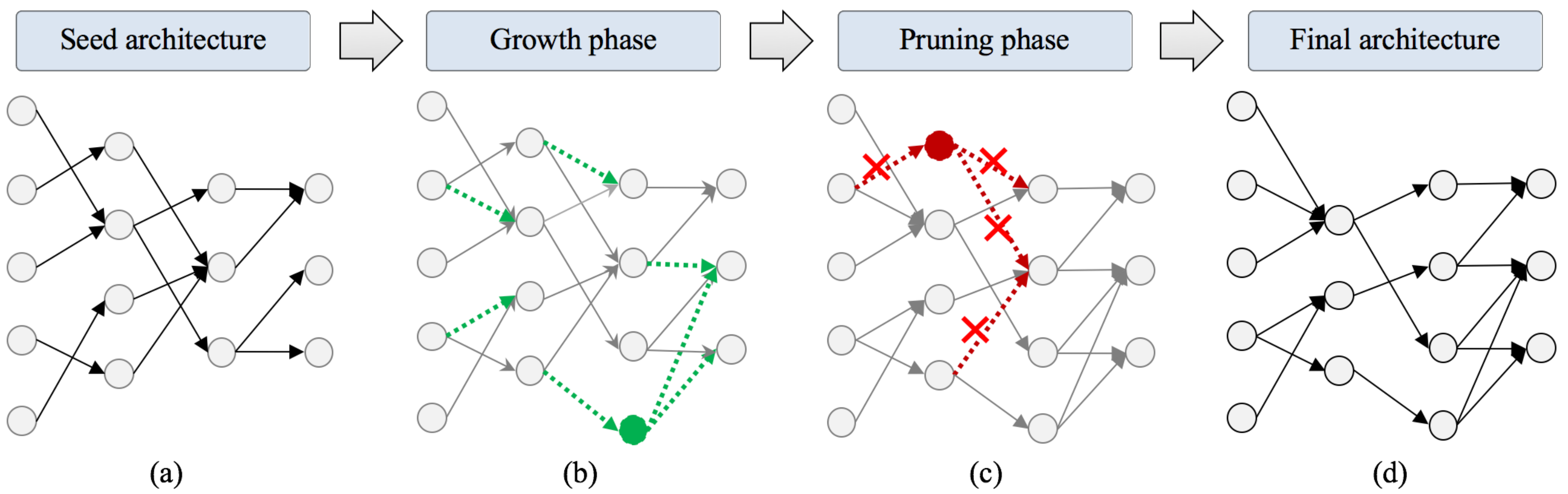}
\end{center}
\caption{An illustration of the grow-and-prune synthesis paradigm used in 
NeST\cite{dai2019nest}.}
\vspace{-2mm}
\label{gnp}
\end{figure}

NeST is a synthesis tool that automatically learns both DNN weights and compact architectures during training\cite{dai2019nest}. The synthesis process contains two phases: (1) a gradient-based growth phase where the gradient information is used to gradually grow new connections, neurons, and feature maps to boost model accuracy, and (2) a magnitude-based pruning phase where redundant connections 
and neurons with low magnitude are removed. The grow-and-prune paradigm in NeST mimics the pattern of synaptic development in the human brain. Specifically, the number of biological synaptic connections increases over the period of a few months after birth and decreases steadily thereafter~\cite{hawkins2017special}.

The NeST flow is illustrated in Fig.~\ref{gnp}. Synthesis starts with a
seed architecture that is initialized as a sparsely-connected DNN but
with all neurons connected. Next, in the growth phase, NeST gradually
grows new connections, neurons, and feature maps based on gradient
information with the goal of increasing accuracy. Then, in the pruning
phase, NeST prunes away redundant and insignificant neurons and
connections based on their magnitudes. After several iterations of
growth and pruning, NeST results in a lightweight DNN model with
drastically reduced size and no accuracy degradation. In addition to its
superior ability to reduce computational cost, NeST offers more 
freedom to DNN designers, since it can start with a wide range of seed architectures and scale to large datasets. For example, for LeNet-300-100 (LeNet-5) on the MNIST dataset, NeST reduces the number of network parameters and floating-point operations (FLOPs) by 70.2$\times$ (74.3$\times$) and 79.4$\times$ (43.7$\times$), respectively. For AlexNet and VGG-16 on the ImageNet dataset, NeST reduces the number of parameters (FLOPs) by 15.7$\times$ (4.6$\times$) and 33.2$\times$ (8.9$\times$), respectively.

\begin{figure}[t]
\begin{center}
\includegraphics[width=\columnwidth]{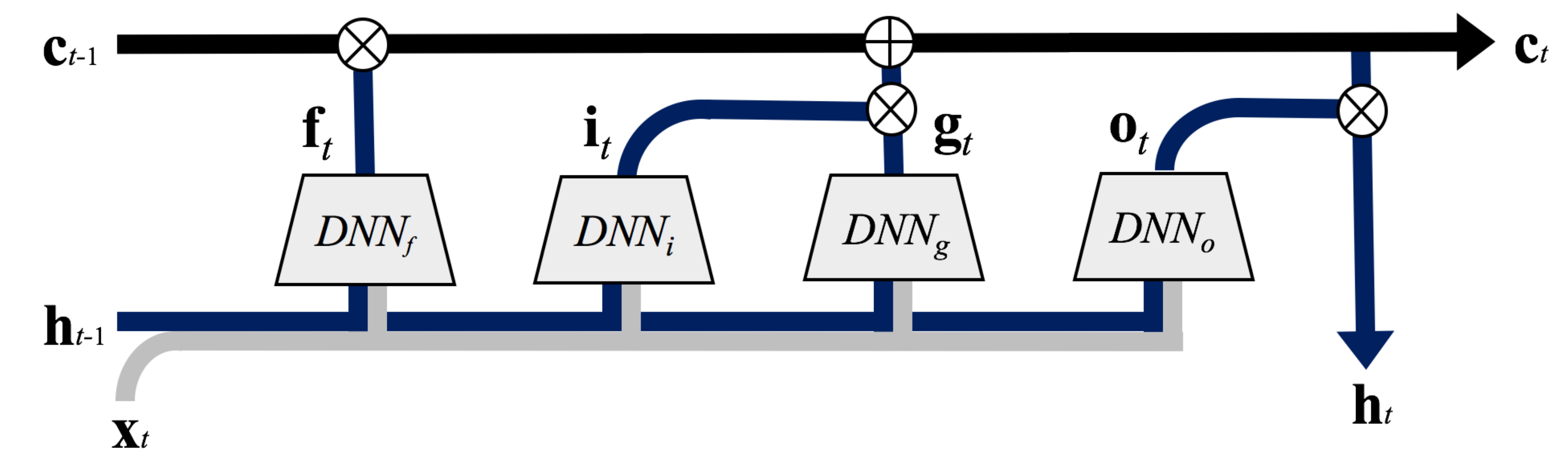}
\end{center}
\caption{Schematic diagram of the H-LSTM structure\cite{dai2018grow}.}
\label{hlstmcell}
\vspace{-3mm}
\end{figure}

Beyond typical feed-forward DNNs, a special type of RNN widely used for sequential data modeling, called LSTM, also faces a model size/accuracy trade-off. To address this problem, a hidden-layer LSTM (H-LSTM) has been proposed~\cite{dai2018grow}. It uses the grow-and-prune paradigm proposed earlier in NeST to synthesize accurate yet compact LSTMs. The structure of an H-LSTM is illustrated in Fig.~\ref{hlstmcell}. As its name suggests, an H-LSTM improves learning by introducing hidden layers in conventional single-layer control gates. As a result, each LSTM cell has greater learning power. Hence, fewer stacking cells are needed to achieve the same or higher accuracy compared to a conventional LSTM. The grow-and-prune method is used to automatically learn the weights and architectures of the control gates. Extensive experiments have been performed to demonstrate the effectiveness of H-LSTMs. Compared to the baseline NeuralTalk architecture, H-LSTMs have 38.7$\times$ fewer parameters (45.5$\times$ fewer FLOPs), 4.5$\times$ lower run-time latency, and 2.8$\%$ higher CIDEr-D score on the MSCOCO dataset. Compared to the DeepSpeech2 architecture on the AN4 dataset, H-LSTMs have 
19.4$\times$ fewer parameters (23.5$\times$ fewer FLOPs) and 37.4$\%$ lower run-time latency. Additionally, H-LSTMs reduce the word error rate from 12.9$\%$ to 8.7$\%$. Compared to the Seq2Seq architecture, H-LSTMs have 10.8$\times$ 
fewer parameters, 14.2$\%$ lower run-time latency, and 3.2$\%$ higher BLEU score on the IWSLT 2014 German-English dataset.

Most automatic architecture synthesis strategies assume a fixed DNN depth during training~\cite{dai2019nest, dai2018grow, molchanov2016pruning,
han2015deep}. By eliminating this constraint, the size of the output model can be further reduced. SCANN\cite{hassantabar2019scann} is a synthesis methodology that generates a general compact feed-forward architecture. In this method, rather than only receiving inputs from the immediate previous layer, hidden neurons can receive inputs from any preceding neuron. In conjunction with dataset dimensionality reduction, layer-wise neural network compression, and global neural network compression, SCANN demonstrates significant model compression power for small- and medium-size datasets: up to 5079$\times$ (geometric mean of 82$\times$) relative to traditional neural networks, with little to no drop in accuracy. This 
can enable energy-efficient inference even on IoT sensors.

\begin{figure}[t]
\begin{center}
\includegraphics[width=\columnwidth]{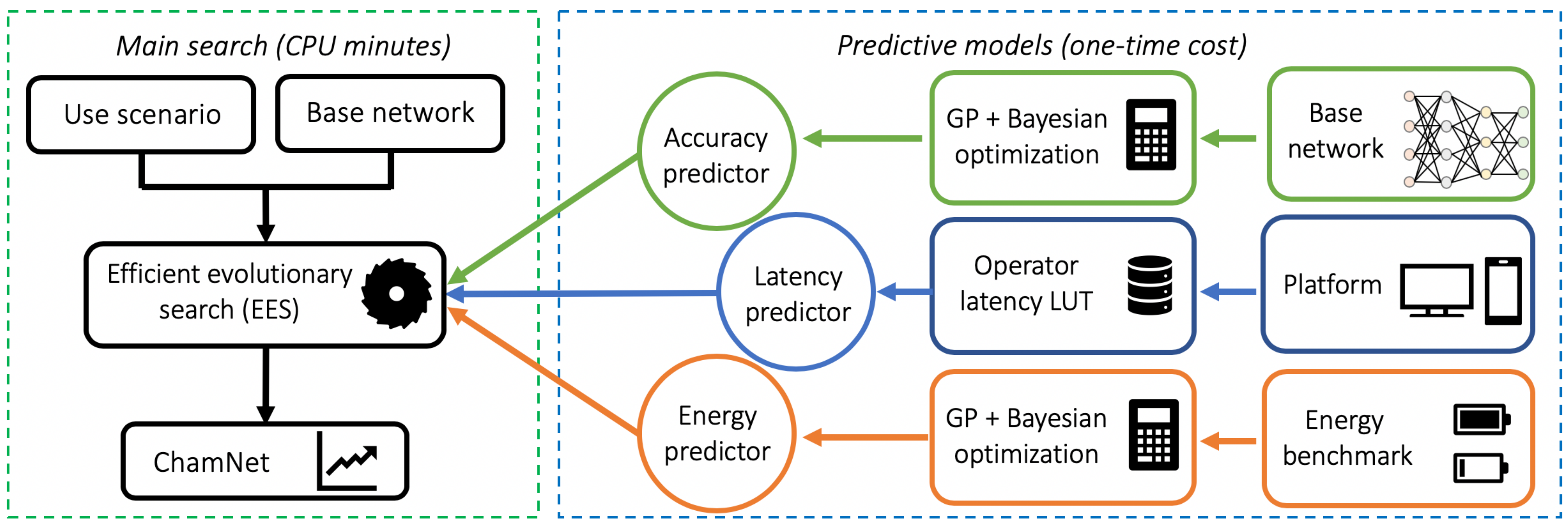}
\end{center}
\caption{The ChamNet model adaptation framework based on efficient 
evolutionary search~\cite{chamnet}.}
\vspace{-3mm}
\label{fig:phases}
\end{figure}

\begin{figure*}[t]
\centerline{\includegraphics[width=\linewidth]{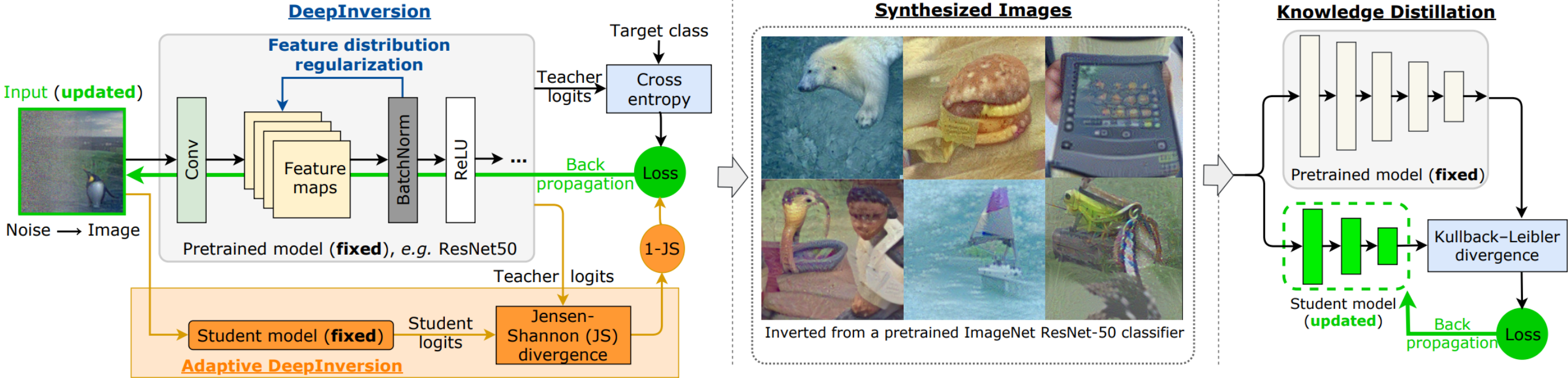}}
\caption{The DeepInversion framework for converting random noise to 
high-fidelity class-conditional images, given just a pretrained CNN. Its 
adaptive version called Adaptive DeepInversion utilizes both the teacher and 
application-dependent student network to improve image diversity. The 
synthesized images can be used to perform various kinds of data-free knowledge 
distillation~\cite{yin2019dreaming}.}
\label{di}
\end{figure*}

\section{Hardware-aware DNN design}
\label{sec:HA}
In this section, we describe an integrated approach that considers hardware specifications and traits, in addition to DNN compactness, to achieve high efficiency and performance.

A large amount of effort in deriving compact and efficient DNNs is directed toward reducing network complexity. However, these approaches are mostly 
hardware-agnostic, and thus complexity reduction is not always tantamount to execution efficiency. To further improve efficiency by considering hardware characteristics, a hardware-guided symbiotic training methodology for execution-efficient and accurate inference models has been proposed~\cite{yin2019hardware}. By leveraging the hardware-impacted hysteresis effect (i.e., the non-monotonic latency surface when the model dimension shrinks) and using a multi-granular grow-and-prune synthesis approach, this method achieves model compactness and accuracy in addition to execution efficiency. For language modeling and speech recognition tasks, the method achieves 7-31$\times$ parameter reduction and 1.7-5.2$\times$ direct latency reduction relative to off-the-shelf NVIDIA GPUs and Intel CPUs, while improving accuracy.

Hardware-guided model adaptation is another new trend for compact DNN
design that honors varying resource budgets upon model deployment. To
enable fast yet automatic model adaptation to the underlying hardware,
the Chameleon framework that leverages existing network building blocks and focuses on exploiting hardware traits to meet target latency and energy constraints has been proposed~\cite{chamnet}. At the core of the approach lie three novel predictors for run-time accuracy, latency, and energy. Based on a one-time profiling effort, these three predictors significantly reduce the intermediate model evaluation time during efficient evolutionary search. In just minutes, the method produces a family of state-of-the-art compact DNNs, called ChamNets, for different targeted platforms with given constraints. For example, it achieves 73.8$\%$ (75.3$\%$) top-1 accuracy on ImageNet with 20ms latency on a mobile CPU (DSP). Compared with MobileNetV2 and MnasNet, ChamNet models achieve up to 8.2$\%$ (4.8$\%$) and 6.7$\%$ (9.3$\%$) absolute top-1 accuracy improvements, respectively, on a mobile CPU (DSP). Compared to ResNet-101 and ResNet-152, ChamNet models achieve 2.7$\%$ (4.6$\%$) and 5.6$\%$ (2.6$\%$) accuracy gains, respectively, on an Nvidia GPU (Intel CPU).

\section{Data Efficiency}
\label{sec:data-efficiency}
Availability of large-scale training data, or a lack thereof, raises another major challenge for DNN deployment. In practice, most datasets are small- or medium-size due to the time and expense of collecting data. In addition, some datasets get updated as new data become available. For example, biomedical datasets may add new patient data over time or include new information as diseases develop. Furthermore, some datasets, such as those containing sensitive information about gender, health, or income, may not be available due to privacy concerns. To maximize efficient data handling, new approaches are being developed that actively exploit information already present in trained models. 

Incorporating new training data effectively and efficiently poses a challenge for model deployment. In typical DNN development, when new data become 
available, the previously trained model is discarded and the DNN is retrained with all the data at hand. This is very inefficient with respect to training 
cost. Ideally, information learned by the neural network from the original data should be preserved, since these data instances still exist in the new 
dataset. A brain-inspired incremental learning framework based on the grow-and-prune synthesis paradigm can be used in this context~\cite{dai2019incremental}. Specifically, when new data arrive, the original model grows new connections in a gradient-directed manner. This
increases the model learning capacity and ability to accommodate new data. Next, the framework prunes away insignificant connections with low-magnitude weights to derive a lightweight yet accurate model. With this incremental learning framework, the cost of training the model with new data is significantly reduced compared to traditional methods, such as retraining the model from scratch and network fine-tuning. In addition, the derived lightweight model is more accurate and requires less inference time. For
ResNet-18 on the ImageNet dataset and DeepSpeech2 on the AN4 dataset, the training cost reductions compared to training from scratch (network fine-tuning) are 64$\%$ (60$\%$) and 67$\%$ (62$\%$), respectively. This method, however, assumes access to the original data when new data arrive.

There is growing concern for protecting data privacy. This concern further challenges efficient DNN development, since data typically used to derive 
compact models may not be available. In these cases, the previously mentioned methods for compact DNN design are not feasible. Recent work has made strides in making efficient DNN derivation data-free with only a trained model. One such method is DeepInversion~\cite{yin2019dreaming}, an image 
synthesis methodology that enables data-free knowledge transfer. One of its many applications is data-free pruning. The DeepInversion framework is 
shown in Fig.~\ref{di}. Given only a trained CNN classifier, DeepInversion can ``invert'' out class-conditional images with high fidelity and diversity, without any prior information about the original dataset. It performs intermediate feature regularization using stored batch norm statistics. This enables generation of high-fidelity instances of all previously seen training classes. To further improve image diversity, an alternative framework, called Adaptive DeepInversion, uses competition regularization to encourage disagreement between a pretrained ``teacher" network and an in-training 
``student" network. The method shows that (1) a trained deep CNN can encode a substantial amount of training distribution information, such that (2) the distribution can be sampled very effectively via DeepInversion. This work demonstrates state-of-the-art performance for data-free knowledge transfer, network pruning, and incremental learning on the CIFAR-10 and ImageNet datasets.

Fundamentally, DeepInversion transforms a discriminative network into a 
generative one. To synthesize images for data-free knowledge transfer, 
DeepInversion performs on par with BigGAN~\cite{brock2018biggan}. Furthermore, 
it does not rely on a model provider training additional 
generative adversarial networks on the original dataset, which can be 
computationally intensive, sensitive to the training setup, and challenging on 
large-scale datasets, e.g., ImageNet1K.

\section{Conclusions and Future Work}
\label{sec:summary}
In this paper, we have reviewed several current trends in compact DNN
design and described synthesis frameworks that automatically generate
lightweight and accurate DNN architectures. So far, most existing work
on efficient DNN synthesis focuses on reducing the computational and storage costs at inference time. However, training DNNs on large-scale datasets remains inefficient and expensive, easily consuming hundreds of GPU hours. More work on speeding up 
training will substantially benefit the entire deep learning community. In addition, more work can be done on novel model design and optimization 
techniques such as hardware-software co-design. The DeepInversion methodology, for example, can be extended to explore different gradient sources for inversion, such as bounding boxes, and can be applied to different DNN architectures, e.g., LSTM and 3D CNN. Extending methods like DeepInversion to other DNN architectures can enable synthesis of video, speech, text, 3D objects, and even language models, while making all these tasks data-free.

\noindent
{\bf Acknowledgments:} This work was supported by NSF under Grant No.
CNS-1907381.

\bibliographystyle{./IEEEtran} 
\bibliography{bib} 

\begin{thebibliography}{10}
\providecommand{\url}[1]{#1}
\csname url@samestyle\endcsname
\providecommand{\newblock}{\relax}
\providecommand{\bibinfo}[2]{#2}
\providecommand{\BIBentrySTDinterwordspacing}{\spaceskip=0pt\relax}
\providecommand{\BIBentryALTinterwordstretchfactor}{4}
\providecommand{\BIBentryALTinterwordspacing}{\spaceskip=\fontdimen2\font plus
\BIBentryALTinterwordstretchfactor\fontdimen3\font minus
  \fontdimen4\font\relax}
\providecommand{\BIBforeignlanguage}[2]{{%
\expandafter\ifx\csname l@#1\endcsname\relax
\typeout{** WARNING: IEEEtran.bst: No hyphenation pattern has been}%
\typeout{** loaded for the language `#1'. Using the pattern for}%
\typeout{** the default language instead.}%
\else
\language=\csname l@#1\endcsname
\fi
#2}}
\providecommand{\BIBdecl}{\relax}
\BIBdecl

\bibitem{human_performance}
Y.~Netzer, T.~Wang, A.~Coates, A.~Bissacco, B.~Wu, and A.~Y. Ng, ``Reading
  digits in natural images with unsupervised feature learning,'' in \emph{Proc.
  NeurIPS}, 2011.

\bibitem{speechlstm}
A.~Graves, A.-R. Mohamed, and G.~E. Hinton, ``Speech recognition with deep
  recurrent neural networks,'' in \emph{Proc. ICASSP}, 2013.

\bibitem{seq2seq}
I.~Sutskever, O.~Vinyals, and Q.~V. Le, ``Sequence to sequence learning with
  neural networks,'' in \emph{Proc. NeurIPS}, 2014.

\bibitem{diabdeep}
H.~Yin, B.~Mukadam, X.~Dai, and N.~K. Jha, ``Diab{D}eep: {P}ervasive diabetes
  diagnosis based on wearable medical sensors and efficient neural networks,''
  \emph{IEEE Trans. Emerging Topics in Computing}, 2019.

\bibitem{he2016deep}
K.~He, X.~Zhang, S.~Ren, and J.~Sun, ``Deep residual learning for image
  recognition,'' in \emph{Proc. CVPR}, 2016.

\bibitem{krizhevsky2012imagenet}
A.~Krizhevsky, I.~Sutskever, and G.~E. Hinton, ``Imagenet classification with
  deep convolutional neural networks,'' in \emph{Proc. NeurIPS}, 2012.

\bibitem{simonyan2014very}
K.~Simonyan and A.~Zisserman, ``Very deep convolutional networks for
  large-scale image recognition,'' in \emph{Proc. ICLR}, 2014.

\bibitem{him}
H.~Yin, Z.~Wang, and N.~K. Jha, ``A hierarchical inference model for
  {Internet-of-Things},'' \emph{IEEE Trans. Multi-Scale Computing Systems},
  vol.~4, pp. 260--271, 2018.

\bibitem{akmandor2018smart}
A.~O. Akmandor, H.~Yin, and N.~K. Jha, ``Smart, secure, yet energy-efficient,
  {I}nternet-of-{T}hings sensors,'' \emph{IEEE Trans. Multi-Scale Computing
  Systems}, vol.~4, no.~4, pp. 914--930, 2018.

\bibitem{wu2018squeezeseg}
B.~Wu, A.~Wan, X.~Yue, and K.~Keutzer, ``Squeeze{S}eg: {C}onvolutional neural
  nets with recurrent {CRF} for real-time road-object segmentation from {3D}
  {LIDAR} point cloud,'' in \emph{Proc. ICRA}, 2018.

\bibitem{yin2017health}
H.~Yin and N.~K. Jha, ``A health decision support system for disease diagnosis
  based on wearable medical sensors and machine learning ensembles,''
  \emph{IEEE Trans. Multi-Scale Computing Systems}, vol.~3, no.~4, pp.
  228--241, 2017.

\bibitem{deepspeech2}
D.~Amodei, S.~Ananthanarayanan, R.~Anubhai, J.~Bai, E.~Battenberg, C.~Case,
  J.~Casper, B.~Catanzaro, Q.~Cheng, G.~Chen \emph{et~al.}, ``{D}eep {S}peech
  2: End-to-end speech recognition in {English} and {Mandarin},'' in
  \emph{Proc. ICML}, 2016.

\bibitem{deepheart}
B.~Ballinger, J.~Hsieh, A.~Singh, N.~Sohoni, J.~Wang, G.~H. Tison, G.~M.
  Marcus, J.~M. Sanchez, C.~Maguire, J.~E. Olgin, and M.~J. Pletcher,
  ``{DeepHeart}: {S}emi-supervised sequence learning for cardiovascular risk
  prediction,'' in \emph{Proc. AAAI}, 2018.

\bibitem{stanford}
Y.~Lin, S.~Han, H.~Mao, Y.~Wang, and W.~J. Dally, ``Deep gradient compression:
  {Reducing} the communication bandwidth for distributed training,'' in
  \emph{Proc. ICLR}, 2018.

\bibitem{wenwei}
W.~Wen, Y.~He, S.~Rajbhandari, W.~Wang, F.~Liu, B.~Hu, Y.~Chen, and H.~Li,
  ``Learning intrinsic sparse structures within long short-term memory,'' in
  \emph{Proc. ICLR}, 2018.

\bibitem{mcculloch1943logical}
W.~S. McCulloch and W.~Pitts, ``A logical calculus of the ideas immanent in
  nervous activity,'' \emph{The {B}ulletin of {M}athematical {B}iophysics},
  vol.~5, no.~4, pp. 115--133, 1943.

\bibitem{rosenblatt1958perceptron}
F.~Rosenblatt, ``The perceptron: {A} probabilistic model for information
  storage and organization in the brain.'' \emph{Psychological {R}eview},
  vol.~65, no.~6, p. 386, 1958.

\bibitem{ivakhnenko1966cybernetic}
A.~G. Ivakhnenko and V.~G. Lapa, ``Cybernetic predicting devices,'' Purdue
  University, Tech. Rep., 1966.

\bibitem{lecun1998gradient}
Y.~LeCun, L.~Bottou, Y.~Bengio, and P.~Haffner, ``Gradient-based learning
  applied to document recognition,'' \emph{Proc. of the IEEE}, vol.~86, no.~11,
  pp. 2278--2324, 1998.

\bibitem{deng2009imagenet}
J.~Deng, W.~Dong, R.~Socher, L.-J. Li, K.~Li, and L.~Fei-Fei, ``Image{N}et: {A}
  large-scale hierarchical image database,'' in \emph{Proc. CVPR}, 2009.

\bibitem{szegedy2015going}
C.~Szegedy, W.~Liu, Y.~Jia, P.~Sermanet, S.~Reed, D.~Anguelov, D.~Erhan,
  V.~Vanhoucke, and A.~Rabinovich, ``Going deeper with convolutions,'' in
  \emph{Proc. CVPR}, 2015.

\bibitem{sandler2018mobilenetv2}
M.~Sandler, A.~Howard, M.~Zhu, A.~Zhmoginov, and L.-C. Chen, ``Mobile{N}et{V}2:
  {I}nverted residuals and linear bottlenecks,'' in \emph{Proc. CVPR}, 2018.

\bibitem{ma2018shufflenet}
N.~Ma, X.~Zhang, H.-T. Zheng, and J.~Sun, ``{ShuffleNet V2}: {Practical}
  guidelines for efficient {CNN} architecture design,'' in \emph{Proc. ECCV},
  2018.

\bibitem{wu2017shift}
B.~Wu, A.~Wan, X.~Yue, P.~Jin, S.~Zhao, N.~Golmant, A.~Gholaminejad,
  J.~Gonzalez, and K.~Keutzer, ``Shift: {A} zero {FLOP}, zero parameter
  alternative to spatial convolutions,'' in \emph{Proc. CVPR}, 2018.

\bibitem{molchanov2016pruning}
P.~Molchanov, S.~Tyree, T.~Karras, T.~Aila, and J.~Kautz, ``Pruning
  convolutional neural networks for resource efficient transfer learning,'' in
  \emph{Proc. ICLR}, 2017.

\bibitem{thinet}
J.-H. Luo, J.~Wu, and W.~Lin, ``Thi{N}et: {A} filter level pruning method for
  deep neural network compression,'' in \emph{Proc. ICCV}, 2017.

\bibitem{nisp}
R.~Yu, A.~Li, C.-F. Chen, J.-H. Lai, V.~I. Morariu, X.~Han, M.~Gao, C.-Y. Lin,
  and L.~S. Davis, ``{NISP}: Pruning networks using neuron importance score
  propagation,'' in \emph{Proc. CVPR}, 2018.

\bibitem{molchanov2019importance}
P.~Molchanov, A.~Mallya, S.~Tyree, I.~Frosio, and J.~Kautz, ``Importance
  estimation for neural network pruning,'' in \emph{Proc. CVPR}, 2019.

\bibitem{han2015deep}
S.~Han, H.~Mao, and W.~J. Dally, ``Deep compression: {C}ompressing deep neural
  networks with pruning, trained quantization and {H}uffman coding,'' in
  \emph{Proc. ICLR}, 2016.

\bibitem{li2016pruning}
H.~Li, A.~Kadav, I.~Durdanovic, H.~Samet, and H.~P. Graf, ``Pruning filters for
  efficient {ConvNets},'' in \emph{Proc. ICLR}, 2017.

\bibitem{liu2017learning}
Z.~Liu, J.~Li, Z.~Shen, G.~Huang, S.~Yan, and C.~Zhang, ``Learning efficient
  convolutional networks through network slimming,'' in \emph{Proc. CVPR},
  2017.

\bibitem{ye2018rethinking}
J.~Ye, X.~Lu, Z.~Lin, and J.~Z. Wang, ``Rethinking the
  smaller-norm-less-informative assumption in channel pruning of convolution
  layers,'' in \emph{Proc. ICLR}, 2018.

\bibitem{zhu2016trained}
C.~Zhu, S.~Han, H.~Mao, and W.~J. Dally, ``Trained ternary quantization,'' in
  \emph{Proc. ICLR}, 2017.

\bibitem{wang2019haq}
K.~Wang, Z.~Liu, Y.~Lin, J.~Lin, and S.~Han, ``{HAQ}: {H}ardware-aware
  automated quantization with mixed precision,'' in \emph{Proc. CVPR}, 2019.

\bibitem{rl3}
B.~Baker, O.~Gupta, N.~Naik, and R.~Raskar, ``Designing neural network
  architectures using reinforcement learning,'' in \emph{Proc. ICLR}, 2017.

\bibitem{nasnet}
B.~Zoph, V.~Vasudevan, J.~Shlens, and Q.~V. Le, ``Learning transferable
  architectures for scalable image recognition,'' in \emph{Proc. CVPR}, 2018.

\bibitem{NASRL}
B.~Zoph and Q.~V. Le, ``Neural architecture search with reinforcement
  learning,'' in \emph{Proc. ICLR}, 2017.

\bibitem{tan2018mnasnet}
M.~Tan, B.~Chen, R.~Pang, V.~Vasudevan, M.~Sandler, A.~Howard, and Q.~V. Le,
  ``{MnasNet}: {P}latform-aware neural architecture search for mobile,'' in
  \emph{Proc. CVPR}, 2019.

\bibitem{fbnet}
B.~Wu, X.~Dai, P.~Zhang, Y.~Wang, F.~Sun, Y.~Wu, Y.~Tian, P.~Vajda, Y.~Jia, and
  K.~Keutzer, ``{FBNet}: Hardware-aware efficient {ConvNet} design via
  differentiable neural architecture search,'' in \emph{Proc. CVPR}, 2019.

\bibitem{neat}
K.~O. Stanley and R.~Miikkulainen, ``Evolving neural networks through
  augmenting topologies,'' \emph{Evolutionary Computation}, vol.~10, no.~2, pp.
  99--127, 2002.

\bibitem{chamnet}
X.~Dai, P.~Zhang, B.~Wu, H.~Yin, F.~Sun, Y.~Wang, M.~Dukhan, Y.~Hu, Y.~Wu,
  Y.~Jia, P.~Vajda, M.~Uyttendaele, and N.~K. Jha, ``{ChamNet}: {Towards}
  efficient network design through platform-aware model adaptation,'' in
  \emph{Proc. CVPR}, 2019.

\bibitem{dai2019nest}
X.~Dai, H.~Yin, and N.~K. Jha, ``{NeST}: {A} neural network synthesis tool
  based on a grow-and-prune paradigm,'' \emph{IEEE Trans. Computers}, vol.~68,
  no.~10, pp. 1487--1497, 2019.

\bibitem{dai2018grow}
------, ``Grow and prune compact, fast, and accurate {LSTMs},'' \emph{IEEE
  Trans. Computers}, vol.~69, pp. 441--452, 2020.

\bibitem{hassantabar2019scann}
S.~Hassantabar, Z.~Wang, and N.~K. Jha, ``{SCANN}: {S}ynthesis of compact and
  accurate neural networks,'' \emph{arXiv preprint arXiv:1904.09090}, 2019.

\bibitem{hawkins2017special}
J.~Hawkins, ``Special report: Can we copy the brain? {W}hat intelligent
  machines need to learn from the neocortex,'' \emph{IEEE Spectrum}, vol.~54,
  no.~6, pp. 34--71, 2017.

\bibitem{yin2019dreaming}
H.~Yin, P.~Molchanov, Z.~Li, J.~M. Alvarez, A.~Mallya, D.~Hoiem, N.~K. Jha, and
  J.~Kautz, ``Dreaming to distill: {D}ata-free knowledge transfer via
  {D}eep{I}nversion,'' in \emph{Proc. CVPR}, 2020.

\bibitem{yin2019hardware}
H.~Yin, G.~Chen, Y.~Li, S.~Che, W.~Zhang, and N.~K. Jha, ``Hardware-guided
  symbiotic training for compact, accurate, yet execution-efficient {LSTM},''
  \emph{arXiv preprint arXiv:1901.10997}, 2019.

\bibitem{dai2019incremental}
X.~Dai, H.~Yin, and N.~K. Jha, ``Incremental learning using a grow-and-prune
  paradigm with efficient neural networks,'' \emph{arXiv preprint
  arXiv:1905.10952}, 2019.

\bibitem{brock2018biggan}
A.~Brock, J.~Donahue, and K.~Simonyan, ``Large scale {GAN} training for high
  fidelity natural image synthesis,'' in \emph{Proc. ICLR}, 2019.

\end{thebibliography}

\end{document}